# A Bayesian Network Classifier that Combines a Finite Mixture Model and a Naïve Bayes Model

**Stefano Monti**[†]
[†]Intelligent Systems Program
University of Pittsburgh
901M CL, Pittsburgh, PA – 15260
smonti@isp.pitt.edu

**Gregory F. Cooper**[†,‡]
[‡]Center for Biomedical Informatics
University of Pittsburgh
8084 Forbes Tower, Pittsburgh, PA – 15213
gfc@cbmi.upmc.edu

## Abstract

In this paper we present a new Bayesian network model for classification that combines the *naive Bayes* (NB) classifier and the *finite mixture* (FM) classifier. The resulting classifier aims at relaxing the strong assumptions on which the two component models are based, in an attempt to improve on their classification performance, both in terms of accuracy and in terms of calibration of the estimated probabilities. The proposed classifier is obtained by superimposing a finite mixture model on the set of feature variables of a naive Bayes classifier. We present experimental results that compare the predictive performance on real datasets of the new classifier with the predictive performance of the NB classifier and the FM classifier.

## 1 Introduction

*When solving a given problem, try to avoid solving a more general problem as an intermediate step.* – Vapnik [32]

The *naive Bayes* (NB) classifier has been the object of active research for several decades. Its use in computer-aided diagnosis, for example, dates back approximately forty years to work by Warner, et al. [33], among others. Despite the considerable advances in probabilistic modeling, the NB model remains to date a very popular classification method, and still is the object of active research [9, 16], as well as a benchmark against which to compare new classifiers. Furthermore, recent results show that the NB model can often outperform more sophisticated classifiers, such as decision trees and Bayesian networks [17], and that its classification performance correlates poorly with the degree of probabilistic dependence among the predictors [9]. In particular, Friedman et al. [17] argue persuasively that generalized learning of BNs might not be appropriate for classification purposes, and show that in many cases it may result in a model with worse predictive performance than the NB classifier. These results can be intuitively understood by noting that learning a BN model for classification corresponds to solving a harder problem than the one we need to solve, namely, the modeling of the joint probability distribution of all the domain variables, rather than the modeling of the conditional distribution of a particular *class* variable given a set of feature variables.

Part of the reason for the NB model's resilience is in the choice of the measure adopted to assess its performance, namely the *classification accuracy* or zero-one loss, defined as the proportion of cases correctly classified by the model (usually measured over a test-set, or cross-validated). Recent results [9, 16] show that a classifier can achieve high classification accuracy even when its probability estimates are poorly calibrated.[1] In a decision-theoretic framework, the probability estimates output by the model are necessary to compute the expected misclassification cost, (i.e., the expected utility) based on which the decision maker would select the *optimal* classification. Poorly calibrated probability estimates may lead to sub-optimal decisions. This is not a problem when the misclassification cost (or utility function) is known and fixed. However, in many circumstances, the misclassification cost is unknown or unequal, that is, case-dependent.[2] In these circumstances, the calibration of the classifier's probability estimates can be as important as its classification accuracy for the assessment of its performance.

In this paper, we provide additional evidence that the NB model is poorly calibrated, and we present a new Bayesian network model for classification that com-bines the *naïve Bayes* (NB) classifier and the *finite mixture* (FM) classifier, in an attempt to improve on the NB classification performance, both in terms of accuracy and in terms of calibration of the estimated probabilities.

---

[1]The probability $p$ of a given classification outcome $c$ is considered perfectly *calibrated* when cases assigned a probability $p$ of yielding outcome $c$, actually yield that outcome a fraction $p$ of the time.

[2]Typical classification problems with unequal misclassification costs are fraud detection [12] and computer-aided medical diagnosis.

The rest of the paper is organized as follows. In Section 2, we briefly describe the BN formalism, and some of the issues involved in learning BNs from data relevant to the classification task. In Section 3, we describe the NB classifier and the FM classifier in some details. In Section 4, we introduce and motivate the new BN classifier. In Section 5 we report the results of our empirical evaluation on real and simulated data. Finally, in Section 6, we summarize the results, and indicate some directions of further research.

## 2   Learning BNs from data

The classifiers that are the focus of this paper are particular instances of Bayesian networks (BNs) [27]. Therefore, it will be advantageous to first introduce those properties of BNs which are common to all three classifiers. Before delving into the topic, we introduce some basic notation.

In general, we denote random variables with upper case letters, such as $X$, $Y$, and their instantiation or realization with the corresponding lower case letters, $x$, $y$, or $x^{(t)}$, $y^{(t)}$, where we use the latter notation when we need to distinguish between different instantiations. Similarly, we denote random vectors with bold upper case letters, such as $V$, $W$, and their instantiation or realization with the corresponding bold lower case letters, $v$, $w$. Given a domain of interest, we denote with $\mathcal{X} = \{X_1, \ldots, X_n\}$ the complete set of variables in that domain. For a discrete variable $X_i$, we denote with $\{x_{i1}, \ldots, x_{ir_i}\}$ its domain of values. We denote with $x$ or $x^{(t)}$ the full instantiations of the variables in $\mathcal{X}$, and with $\mathcal{D} = \{x^{(1)}, \ldots, x^{(N)}\}$, a database of cases over $\mathcal{X}$.

A Bayesian network for the domain $\mathcal{X}$ is defined by a directed-acyclic-graph $M$ over $\mathcal{X}$, and by a set of local probability functions $p(X_i \mid \mathbf{Pa}_i, \theta_i, M)$, each specifying the conditional probability distribution of a variable $X_i$ given its *parent set* $\mathbf{Pa}_i$ (i.e., the set of its immediate predecessors in $M$), with $\theta_i$ denoting the set of parameters necessary to fully characterize the distribution (when there is no ambiguity, we drop the model $M$ in $p(X_i \mid \mathbf{Pa}_i, \theta_i, M)$). A BN allows for the representation of the joint probability of any realization $x$ of all the variables in $\mathcal{X}$ in terms of the local probability functions just defined. That is,

$$p(x \mid \Theta, M) = \prod_i p(x_i \mid \mathbf{pa}_i, \theta_i). \tag{1}$$

In this paper, we consider two possible configurations for the local probability function $p(X_i \mid \mathbf{Pa}_i, \theta_i)$, both restricted to having a parent set $\mathbf{Pa}_i$ containing discrete variables only. When the variable $X_i$ is discrete, we model its conditional probability as a multinomial distribution with a Dirichlet prior over the parameter set $\theta_i$. That is, for each instantiation of the parents $\mathbf{Pa}_i = \mathbf{pa}_{ij}$, the conditional probability of interest has the following specification

$$p(X_i \mid \mathbf{pa}_{ij}, \theta_i) = \text{multinomial}(\theta_{ij}),$$
$$\theta_{ij} \equiv \{\theta_{ij1}, \theta_{ij2}, \ldots, \theta_{ijr_i}\}, \tag{2}$$
$$p(\theta_{ij}) = \text{Dir}(\alpha_{ij1}, \alpha_{ij2}, \ldots, \alpha_{ijr_i}),$$

with each of the $\theta_{ijk}$ representing the probability $p(X_i = x_{ik} \mid \mathbf{Pa}_i = \mathbf{pa}_{ij})$.

When the variable $X_i$ is continuous, we model its conditional probability as a Gaussian distribution with a non-informative prior over its parameter set $\theta_i$. That is, for each instantiation $\mathbf{Pa}_i = \mathbf{pa}_{ij}$,

$$p(X_i \mid \mathbf{pa}_{ij}, \theta_i) = N(\mu_{ij}, \sigma_{ij}^2),$$
$$p(\mu_{ij}, \sigma_{ij}^2) \propto 1/\sigma_{ij}^2. \tag{3}$$

The specification of a prior distribution for the parameters $\theta_i$ given in Equations (2-3) is needed for the purpose of their estimation from data according to the Bayesian paradigm, a point to which we will return.

A well accepted method for learning BNs from data, which is also the one we adopt in this paper, is *model selection*, whereby an attempt is made to select the "best" model (i.e., the best BN structure) according to some properly defined measure – a *scoring metric* – of how well the model fits the data (and possibly our prior beliefs). Once a given BN structure is selected, its parameters need to be estimated.

### 2.1   Model selection

In a Bayesian setting, an appropriate scoring metric for model selection is the posterior probability of the model given the data, $p(M \mid \mathcal{D})$. Based on the proportionality $p(M \mid \mathcal{D}) \propto p(M, \mathcal{D})$, for the purpose of model selection it suffices to compute $p(M, \mathcal{D}) = p(\mathcal{D} \mid M)p(M)$. The term $p(M)$ is the prior probability of the model, and it needs to be provided as input. The term $p(\mathcal{D} \mid M)$ is usually called the *marginal likelihood*, or *integrated likelihood*, or *evidence*, and its computation involves the solution of a high dimensional integral.

When there are missing values or hidden variables, the computation of this integral is in general computationally infeasible, and approximate methods must be used. A class of approximations which is widely used is the class of asymptotic approximations [21]. Many of these approximations can be interpreted as different versions of *penalized likelihood*, since they have the general form

$$p(\mathcal{D} \mid M) \simeq p(\mathcal{D} \mid \hat{\Theta}, M) + penalty, \qquad (4)$$

where $\hat{\Theta}$ is the maximum likelihood (ML) estimator of the parameters $\Theta$ (which can also be replaced by the mode $\hat{\Theta}$), and *penalty* can in general be interpreted as a term that penalizes model complexity, and its exact form differs in the various versions of penalized likelihood proposed. This class of approximations includes the *Bayesian Information Criterion* (BIC) [31], the *Akaike Information Criterion* (AIC), the score developed for AutoClass [4], usually referred to as the *Cheeseman-Stutz* (CS) approximation, and the Integrated classification likelihood (ICL) [2].[3] These are also the scores that we use in the experimental evaluation described in Section 5. All these approximations rely on the availability of the ML estimator $\hat{\Theta}$, or the mode $\hat{\Theta}$, of the parameter $\Theta$. We turn to the computation of these quantities next.

## 2.2 Parameter estimation

In a Bayesian setting, the estimation of the parameters $\theta_i$ in $p(X_i \mid \mathbf{Pa}_i, \theta_i)$ usually consists of finding the *maximum a posteriori* (MAP) configuration, or mode, of $\theta_i$ [1]. Given a complete dataset $\mathcal{D}$ (i.e., a dataset with no missing values), for a discrete variable $X_i$ with multinomial distribution as specified in Equation (2), the MAP estimation of the parameters in $\theta_i$ yields

$$\hat{\theta}_{ijk} = \frac{\alpha_{ijk} + N_{ijk}}{\alpha_{ij} + N_{ij}}, \qquad (5)$$

where $N_{ijk}$ and $N_{ij}$ are the *sufficient statistics* of the data, with $N_{ijk}$ denoting the number of cases in the database $\mathcal{D}$ where $X_i = x_{ik} \wedge \mathbf{Pa}_i = \mathbf{pa}_{ij}$, and $N_{ij} = \sum_k N_{ijk}$ (when $X_i$ has no parents, clearly $N_{ij}$ reduces to $N$).[4] In a more formal notation, which will become useful later, $N_{ijk}$ can be computed as follows

$$N_{ijk} = \sum_{l=1}^{N} 1_{\{x_i^{(l)} = x_{ik} \wedge \mathbf{pa}_i^{(l)} = \mathbf{pa}_{ij}\}}, \qquad (6)$$

where $1_{\{\cdot\}}$ is the indication function (i.e., $1_{\{cond\}} = 1$, if *cond* holds, 0 otherwise).

For a continuous variable with Gaussian distribution as specified in Equation (3), the MAP estimation of

---

[3]For an extensive analysis and comparison of some of these approximations, we refer the interested readers to [5].

[4]If we set all $\alpha_{ijk}$ to 1, we obtain the well know Laplace's rule of succession [1]. In general the use of Bayesian estimates in place of maximum likelihood (ML) estimates can be interpreted as a form of *smoothing*, since their most notable effect is to give a minimal positive probability to those estimates that would be 0 according to ML.

the parameters in $\theta_i$ yields

$$\tilde{u}_{ij} = \frac{1}{N_{ij}} \sum_{l=1}^{N} x_i^{(l)} 1_{\{\mathbf{pa}_i^{(l)} = \mathbf{pa}_{ij}\}},$$

$$\tilde{\sigma}_{ij}^2 = \frac{N_{ij} - 1}{N_{ij}(N_{ij} - 3)} \sum_{l=1}^{N} (x_i^{(l)} - \tilde{u}_{ij})^2 1_{\{\mathbf{pa}_i^{(l)} = \mathbf{pa}_{ij}\}}. \qquad (7)$$

In other words, we take $(\mu_{ij}, \sigma_{ij}^2)$ to be the mean and variance of $X_i$ over the subset of cases in $\mathcal{D}$ where the parent set $\mathbf{Pa}_i$ takes its $j$-th value (see, e.g., [1] for the details on the Bayesian estimates).

## 2.3 Parameter estimation with missing values

When the dataset $\mathcal{D}$ contains missing values and/or hidden variables, in most cases exact parameter estimation is not computationally feasible, and approximate methods need to be used. The approximate method we adopt is the EM algorithm [8, 24], a technique particularly suitable for those maximization problems that can be characterized as "missing data", or "latent variables" problems.

EM is an iterative procedure which, starting from some initial (possibly random) parameterization of the model, fills in the missing data with their expected value according to the current parameterization (*expectation* step), and uses the filled-in data to newly estimate the parameters (*maximization* step), repeating these two steps until convergence. The important feature of the EM algorithm is that, save a few degenerate cases (e.g., see [24], Section 3.6), it is guaranteed to converge to a (possibly local) maximum.

EM's relative efficiency is due to the fact that the algorithm does not actually estimate the missing values individually. Rather, it works with those functions of the missing values necessary to estimate the model parameters. For example, when dealing with distributions from the exponential family, EM works with the expected sufficient statistics.

With reference to the conditional multinomial probability distribution of Equation (2), the relevant sufficient statistics are $N_{ijk}$ and $N_{ij}$, as used in Equation (5). The computation of their expectation according to the distribution $p(\cdot \mid \Theta^{\text{old}}, M)$ yields

$$\mathrm{E}[N_{ijk} \mid \mathcal{D}, \Theta^{\text{old}}] = \sum_{l=1}^{N} p(x_{ik}, \mathbf{pa}_{ij} \mid x^{(l)}, \Theta^{\text{old}}), \qquad (8)$$

and $\mathrm{E}[N_{ij} \mid \mathcal{D}, \Theta^{\text{old}}] = \sum_k \mathrm{E}[N_{ijk} \mid \mathcal{D}, \Theta^{\text{old}}]$. These expected sufficient statistics can then be substituted in Equation (5) to compute the new value of $\Theta^{\text{new}}$.

For the conditional Gaussian probability distribution of Equation (3), the estimation is similarly modified in the case of missing data, and it is obtained by replacing in Equation (7) the sufficient statistic $N_{ij}$ with its expectation computed according to Equation (8), and by substituting the indicator function $1_{\{\mathbf{pa}_i^{(l)}=\mathbf{pa}_{ij}\}}$ with the probability $p(\mathbf{pa}_{ij} \,|\, x^{(l)}, \Theta^{\mathrm{old}})$.

It is worth noting the similarity between Equation (6) and Equation (8). The latter is in fact obtained from the former by replacing the *hard* partition determined by the indicator function of Equation (6), with the *soft* partition determined by the posterior probability of a given configuration of Equation (8). That is, in Equation (8) a fractional assignment is performed, such that the $l$-th case contributes to the count $N_{ijk}$ a fraction equal to the posterior probability $p(x_{ik}, \mathbf{pa}_{ij} \,|\, x^{(l)}, \Theta^{\mathrm{old}})$.

This observation also suggests a straightforward modification of EM applicable when the variables with missing values are discrete. This modification is obtained by computing the expected sufficient statistics as

$$\mathrm{E}[N_{ijk}|\mathcal{D}, \Theta^{\mathrm{old}}] = \sum_{l=1}^{N} \tilde{z}_{ijk}^{(l)}, \qquad (9)$$

with each of the $\tilde{z}_{ijk}^{(l)}$ defined as

$$\tilde{z}_{ijk}^{(l)} = \begin{cases} 1 & \text{if } jk = \mathrm{argmax}_{j'k'}[p(x_{ik'}, \mathbf{pa}_{ij'} \,|\, x^{(l)})] \\ 0 & \text{otherwise.} \end{cases}$$

That is, no fractional assignments are made, and each case with missing values is filled in according to the configuration maximizing $p(\cdot \,|\, x^{(l)}, \Theta^{\mathrm{old}})$

This variant of EM is known as the *classification* EM (CEM) algorithm [3], and it can be thought of as an attempt at finding the parameterization of a model that best partitions the data. For this reason, it is particularly appropriate for use in probabilistic clustering based on the FM model, to be discussed in the next section, and it is an integral part of the ICL approximation of the marginal likelihood, $p(\mathcal{D} \,|\, M)$, introduced in [2] and used in our experimental evaluation.

## 3    Bayesian network classifiers

In this section, we briefly review the BN classifiers that are the building blocks of the new classifier to be introduced in the next section. We will consider the classification domain $\mathcal{X} = \{C\} \cup \boldsymbol{X}$, where $C$ is a class variable taking values in the domain $\{c_1, c_2, \dots, c_{r_c}\}$, and $\boldsymbol{X} = \{X_1, X_2, \dots, X_n\}$ is a set of feature variables that can be either discrete or continuous.

The common feature of all the BN classifiers we are going to describe is in the fact that the probability

$p(C \,|\, \boldsymbol{X}, \Theta, M)$ of the class variable $C$ conditioned on the set of feature variables $\boldsymbol{X}$ can be computed very efficiently, due to the assumptions of conditional independence encoded in the classifiers' structure.

### 3.1    The naïve-Bayes model

The Naïve-Bayes model (NB) [11], an example of which is shown in Figure 1.a, is based on the assumption that the feature variables are independent given the class variable. This assumption is often unrealistic, but it allows for a very parsimonious representation of the joint probability over the variables of interest,

$$p(C, \boldsymbol{X}) = p(C) \prod_{i=1}^{n} p(X_i \,|\, C). \qquad (10)$$

From this factorization, it follows that for the specification of the model, we need to specify the prior probability $p(C)$, and the probabilities $p(X_i \,|\, C)$ of each feature variable $X_i$ conditioned on the class variable.

It also follows that learning a NB model from data consists of estimating these probabilities, which can be easily done as illustrated in the previous section.[5]

While the NB model allows for the direct modeling of continuous variables, empirical results show that we can often obtain better classification performance if the continuous feature variables are discretized [10]. This suggests that imposing a given parametric form on the distribution of the continuous variables is more restrictive than treating these variables as discrete, a point to which we will return.

### 3.2    The finite mixture model

A finite mixture model (FM), as shown in Figure 1.b, has the same structure as a NB model. However, in the FM model the class variable is itself a child node, and the common parent is a hidden variable. The parent node $H$ represents an unmeasured, discrete variable, and it is used to model the interaction among the feature variables $\{X_i\}$, as well as among the feature variables and the class variable $C$. Notice that the FM model can represent a much wider set of probability distributions than the NB model, since it imposes less constraints on the dependency structure. However, because of the FM model's structure, the class variable loses the "preferential" status it was accorded in the NB model. This suggests that the FM model might in general be a better approximator of the joint probability over the observable variables than the corresponding NB model, but this could come at the expense of its predictive accuracy with respect to the class variable. In a FM model, the joint probability distribution

---

[5]We are assuming that no feature selection is performed.

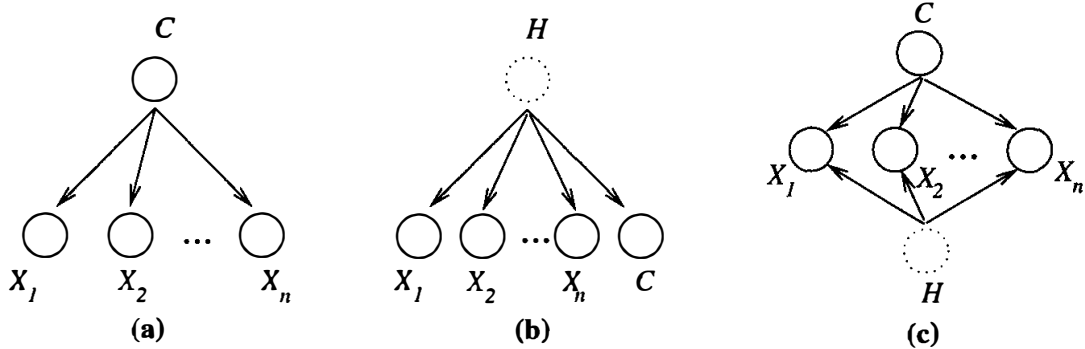

Figure 1: a) The Naïve-Bayes (NB) model, with classification variable $C$, and set of feature variables $\{X_i\}$, which are independent given $C$. b) The finite mixture (FM) model, where the hidden variable $H$ models the dependencies among the feature variables $\{X_i\}$, and among the feature variables and the class variable $C$. c) The finite-mixture-augmented naïve-Bayes (FAN) model, obtained by superimposing an FM model on the set of feature variables of an NB model.

of the observable variables can be computed as follows:

$$p(C, \boldsymbol{X}) = \sum_{k=1}^{r_h} \left[ p(h_k)p(C \mid h_k)\prod_{i=1}^{n} p(X_i \mid h_k) \right], \quad (11)$$

where $r_h$ denotes the number of *mixture components* (i.e., the number of values of the hidden variable), and where the $p(h_k)$'s are the *mixing proportions*.

Kontkanen et al. [23] have obtained good results using this model for classification. Notice that, as in the NB model, the feature variables can be either continuous or discrete. Unlike in the NB model, however, in the FM model the variable $H$ is hidden, and its number of values is unknown. Therefore, exact estimation techniques are not applicable.

Learning a FM model from data thus involves: 1) the determination of its number of components (*models selection*)[6]; and 2) given the number of components, the estimation of the models' parameters.

The parameter estimation is a straightforward instantiation of the general procedure based on the EM algorithm described in the previous section. The model selection is the most difficult step. As explained in Section 2, in a Bayesian setting an appropriate scoring metric for model selection is the joint probability $p(M, \mathcal{D})$ of a model $M$ and data $\mathcal{D}$, which reduces to the marginal likelihood $p(\mathcal{D} \mid M)$ under the assumption of an uninformative prior distribution over the models' space. We will use this scoring metric to select the proper number of components of a FM model, and we will make use of the asymptotic approximations described in Section 2.1, for its computation.

<hr/>

[6]In general, the selection of a FM model would also involve the choice of the parametric form of the probability distributions used in the model. However, as explained in Section 2, we restrict this form to a single choice, thus eliminating this dimension from the model selection.

## 4   A new BN classifier

The classifier that we describe in this section combines the two models described in the previous section, while relaxing the assumptions on which they are based. We call the new classifier the *finite-mixture-augmented Naïve Bayes (FAN) model*, as it is obtained by superimposing a finite mixture model on the set of feature variables of a naive Bayes model. An example of the FAN model is depicted in Figure 1.c. The hidden variable $H$ is introduced to model the residual probabilistic dependencies among the feature variables $\{X_i\}$ that are not captured by the class variable $C$. At the same time, in an attempt to improve over the FM model, the FAN model reduces the burden on the hidden variable $H$ by modeling part of the dependencies among feature variables through the class variable $C$.

The dependency structure of the FAN model allows for the following factorization of the joint probability of the observable

$$p(C, \boldsymbol{X}) = p(C)\sum_{k=1}^{r_h} \left[ p(h_k)\prod_{i=1}^{n} p(X_i \mid C, h_k) \right], \quad (12)$$

where, as before, $r_k$ denotes the number of the model's component, and the $p(h_k)$ are the mixing proportions.

The parameter estimation for the FAN model is a straightforward adaptation of the corresponding procedure for the FM model, the only difference due to the additional common parent $C$. From the factorization of Equation (12), it follows that we need to estimate the conditional probabilities $p(X_i \mid C, H)$ for each feature variable $X_i$, the mixing proportions $p(H)$, and the prior probability $p(C)$. These can all be estimated as described in Section 2. The selection of the number of mixture components is identical to the corresponding task for the induction of a FM model.

Notice that the FAN model subsumes NB, in that the latter corresponds to a a FAN model with a one-valued hidden variable $H$. Furthermore, while FAN is a more general model than NB, it conserves of the latter the preferential status accorded to the class variable.

With respect to the FM model, since the FAN model encodes part of the dependencies among feature variables through the class variable $C$, we would expect it to have a smaller number of components than the FM model for the corresponding domain (as it is confirmed by our experimental evaluation). This is a desirable consequence, since the rate of convergence of the EM algorithm, which we use for parameter estimation, is a function of the ratio of the observed information (provided by the observed data) to the missing information (provided by the missing data), a property usually referred to as the *missing information principle* (e.g., see [24], Section 3.8). Therefore, the reduced number of components might well translate into faster convergence of the parameter estimation task, as well as of the model selection task.

### 4.1   Related work

The literature on methods for generalizing the NB classifier is quite extensive. Recent examples are different versions of *boosted* NB [15, 29], and the decision-tree-NB hybrid introduced in [22]. However, a review of the relevant literature is beyond the scope of this paper. In this section, we focus only on two specific approaches to Bayesian classification because of their considerable similarity to ours.

The FAN model is similar in spirit to the method of discriminant analysis by Gaussian mixtures (MDA) [20]. However, MDA as defined in [20] is restricted to domains containing continuous features only, with the distribution of each feature modeled as a Gaussian distribution with independent mean and common variance/covariance matrix. Furthermore, in MDA, a mixture model with a (possibly) different number of components is specified for each of the class values. Finally, in MDA no model selection is involved, since the number of mixture components needs to be supplied as input.

A BN classifier that shares our goal to relax the NB's assumption of conditional independence is the *Tree Augmented Naive Bayes (TAN) model* described in [17]. TAN is an extension of the NB model that uses a poly-tree structure defined over the set of feature variables to model the dependencies among feature variables not captured by the class variable. While the use of the TAN model as described in [17] is restricted to domains containing discrete variables only, an extension of the model that allows for the inclusion of continuous features is described in [18].

## 5   Experimental evaluation

This section illustrates our evaluation methodology and reports the empirical results we obtained. We performed experiments on both simulated and real data. We start this section with a brief discussion of the summary statistics we use for model assessment and comparison. We then describe the experiments with simulated data. Finally we report the results of the experiments with real data.

### 5.1   Summary statistics

The summary statistics we use are: 1) the *classification accuracy*, defined as the proportion of cases correctly classified; 2) the area under the *ROC curve* (for binary classification only), defined below; 3) the *empirical conditional entropy* (C̃E), defined below; and 4) in the experiments with simulated data, the *structural difference*, defined as the difference in the number of mixture components between the learned model and the "true" (i.e., the generating) model.

We define the empirical conditional entropy over a test set $\mathcal{D}_{\text{test}}$ of size $L$ as

$$\tilde{\text{CE}}(\mathcal{D}_{\text{test}}) = -\frac{1}{L}\sum_{l=1}^{L}\log\tilde{p}(c^{(l)}\,|\,\boldsymbol{x}^{(l)},\boldsymbol{\Theta},M)\,,\qquad(13)$$

to be used as a measure of a classifier's calibration. In fact, C̃E is a Monte Carlo approximation of the conditional entropy $-\sum_{C,\boldsymbol{X}}p(C,\boldsymbol{X})\log\tilde{p}(C\,|\,\boldsymbol{X})$, where $p$ is the distribution according to which the data is generated, and $\tilde{p}$ is the probability estimator provided by the classifier. It can be shown that the conditional entropy is minimized when $\tilde{p}=p$ [6]. The C̃E can be interpreted as a measure of the distance between the "true" distribution and the classifier's distribution.

The area under the ROC (receiver operating characteristic) curve [19] is becoming a well accepted alternative to classification accuracy as a measure of classification performance in domains with a binary class variable [28]. The ROC curve plots the true positive rate as a function of the false positive rate as we vary from 0 to 1 the threshold that we use to decide whether to classify a given case as positive or negative. The area under the curve can be used as a measure of classification performance. An area of 1 corresponds to perfect accuracy. An area of 0.5 corresponds to the performance of a classifier that randomly guesses the outcome. Since the ROC curve can only be computed for domains with a binary class variables, we report this measure only for a subset of the databases used in the experiments.

### 5.1.1   Evaluation with simulated data

The experiments described in this section are aimed at evaluating the appropriateness of the asymptotic approximations defined in Section 2.1 for the purpose of selecting a FAN classifier. The use of simulated data based on a known generating model allows us to compare the learned model to the generating one. In particular, it allows us to determine to what extent the number of mixture components of the learned model – and how this differs from the correct number of components – affect classification performance.

The synthetic models used for this experiment were created as follows. We selected a small subset of databases (DBs) from the UCI repository [25] (3 DBs with discrete features, 3 DBs with continuous features, and 2 DBs with both discrete and continuous features). For each of these DBs, we induced a FAN model (based on the BIC score) and we used it as the gold standard (GS). The parameterization of the GSs based on real data is aimed at maximizing the plausibility of the resulting models. Based on each GS, and for three distinct sample sizes, we generated 10 distinct train-set/testset pairs from the GS (the size of the testset is the same irrespective of the trainset size). FAN models were induced from the trainsets (for each of the asymptotic approximations: BIC, AIC, CS, and ICL), and summary statistics over the testsets were analyzed.

The results of these experiments can be summarized as follows. The classifiers induced based on the ICL score performed best with respect to every summary statistic (accuracy, ROC area, cross-entropy and structural difference). BIC was second best with respect to accuracy, cross-entropy and structural difference. Unexpectedly, given the results of previous studies [5], CS was second best with respect to the ROC area only. AIC always performed the worst.

We also tested for correlation between each of the classification performance measures (accuracy, ROC area, CE) and the structural difference (we used the Spearman's Rho rank correlation for this test). That is, we tested whether selecting the correct number of mixture components would positively affect the classification performance. As would be expected, the experiments revealed a significant negative correlation ($p \ll .01$) between each of the performance measures and the structural difference, with the index of correlation ranging between $-0.45$ and $-0.55$.

### 5.2   Evaluation with real data

For the evaluation with real data, we used a sample of DBs from the repository at UC Irvine. We also used the PORT database [14], a medical DB of pneumonia patients gathered by clinical researchers at the University of Pittsburgh, and several other medical centers. We used a total of 56 DBs. The classification performance measures were obtained by 10-fold stratified cross-validation for 42 DBs, and by hold-out for the remaining 14 large-size DBs, with the trainset/testset partition as specified by the DB donors.[7]

For the DBs containing both continuous and discrete variables, we report classification results based on both discretized and non-discretized data. For the discretization of the continuous variables, we used our implementation of the class-based discretization method described in [13]. The total of 56 DBs is obtained by counting separately the discretized and non-discretized version of each DB.

Since ICL is the approximation that performed best on the simulated data, the results on real data are those obtained based on this approximation.[8] For lack of space, we mainly focus our attention on the comparison between the FAN model and the NB model.

**Classification accuracy**: Figure 2 (left) plots the accuracy of the FAN model against the accuracy of the NB model for each DB. Overall, FAN performed better than NB on 23 DBs, while NB outperformed FAN on 14 DBs (as it is shown below however, not all of these differences are statistically significant). For brevity, we denote this result as a 23/14 FAN *vs* NB ratio, and we will use this notation in the following comparisons. In the remaining 19 DBs the induced FAN model had only one component, thus reducing to NB. If we test for the statical significance of the difference in classification accuracy (by a binomial test, or Mcnemar test as described in [30]), we obtain a 14/1 FAN *vs* NB ratio at the 95% confidence level (i.e., FAN outperformed NB in 14 DBs, and NB outperformed FAN in 1 DB, with all differences having a p-value $p \leq .05$), and a 14/0 FAN *vs* NB ratio at the 99% confidence level ($p \leq .01$). Although not shown, both FAN and NB performed significantly better than FM in terms of classification accuracy, with a 25/0 FAN *vs* FM ratio, and a 21/6 NB *vs* FM ratio, both at the 99% confidence level.

Notice that the significance levels just reported do not take into account the effect of multiple testing. For this reason, we also performed tests *across* DBs, aimed at testing the probability of obtaining the observed differences across DBs by chance (e.g., when comparing FAN and NB with respect to their classification accuracy, the null hypothesis would correspond to having

---

[7]For the full list of DBs used, and for the complete tables reporting classification accuracy, empirical CE, and ROC curve area, for each DB, see [26].

[8]Although not reported here, the results and conclusions obtained based on the other approximations are largely in agreement with those obtained based on ICL.

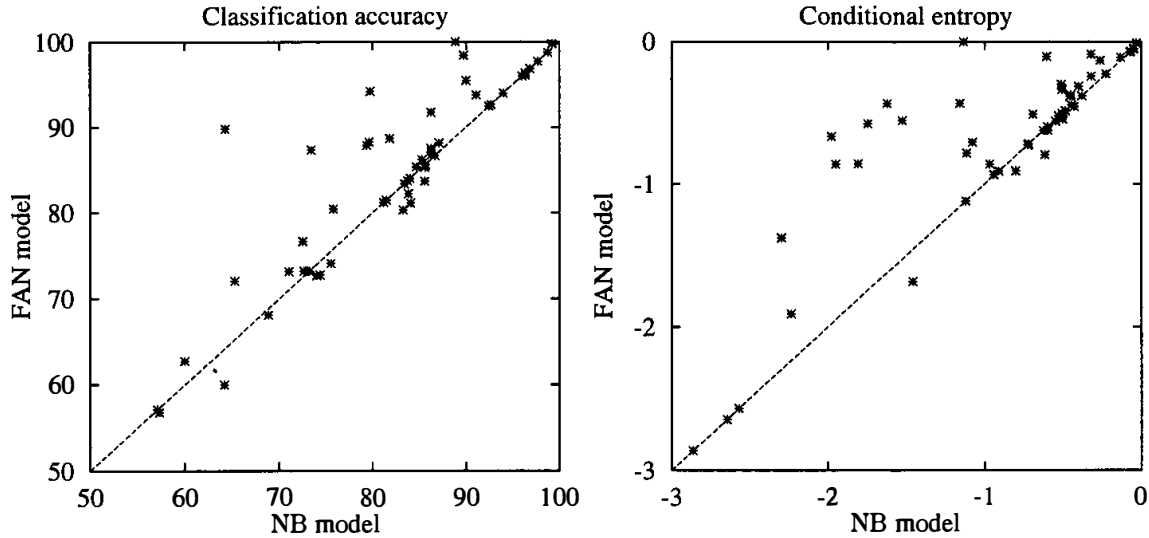

Figure 2: Comparison of the classification performance of the NB model and the FAN model, in terms of classification accuracy (left plot), and in terms of calibration (right plot). Data points above the diagonal line favor the FAN model.

FAN and NB perform the same on all DBs or, equivalently, having FAN perform better than NB in about half of the DBs, and having NB perform better than FAN in the other half). Also, because we used DBs of unequal size, it may happen that for a large DB a small difference in classification accuracy is statistically significant, while for a small DB a large difference in the performance measure is not statistically significant. Therefore, we performed two tests, one aimed at testing for the *magnitude* of the differences, and the other aimed at testing for the *significance* of the differences. In the former, we measured the difference of the classification accuracy for the two classifiers to be compared for each DB, and we tested whether the median of these differences was significantly different from 0. In the latter, for each DB we recorded the $z$-score of the classification accuracy (i.e., we recorded the test statistic used to assess the significance of the difference between the two classifiers being compared on the given DB) and we tested whether the median of these $z$-scores was significantly different from 0. In both tests we used a signed-rank test. Table 1 illus-trates the design of the two tests. Table 2 reports the p-values of the pairwise comparisons. All the observed differences are statistically significant, with $p \ll 0.01$.

|  | p-value of magnitude | p-value of significance |
|---|---|---|
| FAN $vs$ NB | 0.008 | 0.0003 |
| FAN $vs$ FM | $\ll$ 1e-5 | $\ll$ 1e-5 |
| NB $vs$ FM | 0.0002 | $\ll$ 1e-5 |

Table 2: p-values of the pairwise comparisons aimed at assessing the magnitude and the significance of the observed difference in the classification accuracy across DBs.

**Calibration:** Figure 2 (right) also plots the negative of the empirical conditional entropy ($\hat{CE}$) of the FAN model against the $\hat{CE}$ of the NB for each DB. Overall, we attained a 28/9 FAN $vs$ NB ratio (as explained above, FAN reduced to NB in the remaining 19 DBs). If we test for the significance of the differences (we used a paired t-test for this comparison), we obtain a 26/4 ratio at the 95% confidence level, and a 24/2 ratio at the 99% confidence level. As expected, the FM model outperformed both FAN and NB in terms of calibration, with a 27/2 FM $vs$ NB ratio, and a 15/8 FM $vs$ FAN ratio, both at the 99% confidence level.

**Area under the ROC curve:** Finally, Figure 3 plots the area under the ROC curve of FAN against that of NB for each DB with a binary class variable (a total of 24 DBs). Overall, we obtained a 8/13 FAN $vs$ NB ratio, which reduced to a 4/7 ratio when considering only differences statistically significant at the 95% confidence level, and to a 2/4 ratio at the 99% confi-dence level. Both FAN and NB outperformed FM as measured by the area under the ROC curve.

| DB | $M_1$ | $M_2$ | magnitude | significance |
|---|---|---|---|---|
| db$_1$ | $a_{11}$ | $a_{21}$ | $d_1 = a_{11} - a_{21}$ | $z_1$ |
| ... | ... | ... |  |  |
| db$_n$ | $a_{1n}$ | $a_{2n}$ | $d_n = a_{1n} - a_{2n}$ | $z_n$ |
| test |  |  | median($d_i$)$\neq 0$ | median($z_i$)$\neq 0$ |

Table 1: Testing the statistical significance of the observed differences across DBs. $M_1$ and $M_2$ denote the two classifiers being compared (e.g., FAN and NB), $a_{ji}$ denotes the classification accuracy of the classifier $M_j$ on the database db$_i$, and $z_i$ indicates the test statistic for db$_i$.

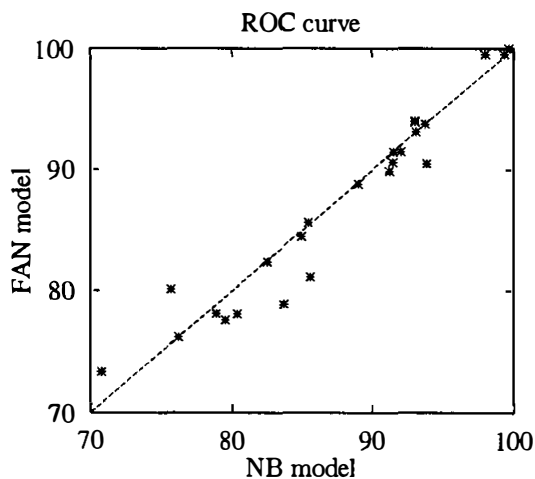

Figure 3: Comparison of NB and FAN, in terms of the area under the ROC curve.

**Time requirements**: both FAN and FM are several order of magnitude slower than the NB model in terms of induction time. In particular, in our experimental evaluation, the FAN model on average needed 200 CPU seconds of training for each CPU second for the NB model, and the FM model needed about 300 CPU seconds for each CPU second for the NB model. Notice however that the computational time for classification (i.e., for classifying a new unseen case) is comparable in the three models. That is, if $T$ is the classification time for NB, than the classification time of FAN is approximately $r_h T$, where $r_h$ is the number of values of the hidden variable. Similarly, the classification time for FM is approximately $\frac{r_c}{2}T$, where $r_c$ is the number of values of the class variable.

### 5.3 Discussion

The results presented provide evidence that the new classifier can often outperform the NB classifier in terms of classification accuracy, while significantly improving the calibration of the probability estimates.

The inconsistency between the empirical results in terms of classification accuracy (which favor FAN) and in terms of ROC curve area (which slightly favor NB) can be explained by noticing that FAN tended to perform better than NB in those DBs with a many-valued class variable, for which the ROC curve area could not be computed. A tentative explanation for the better performance of FAN over NB on DBs with a multi-valued class variable can again be found in the poor calibration of NB. When the class variable has many values, the poorly calibrated probability estimates of

NB are more likely to lead to the wrong classification.

From the results, it also appears that the FAN model performed poorly on the DBs containing continuous variables. In 10 DBs containing continuous variables, NB outperformed FAN. However, when using discretized data, in 8 out of those 10 DBs FAN outperformed NB. We believe that the main reason for this result is in our use of univariate Gaussians (with a distinct standard deviation for each value of the conditioning parent set of the continuous variable) for modeling the conditional probability of continuous variables. As noted in [7], when some component standard deviations become very small relative to others (usually generated by a small number of data points sufficiently close to each other), this can cause the EM algorithm to select a spurious local maximizer, thus leading to poor parameter estimation, and consequently to poor model selection.

Finally, the FAN model performed consistently better than the NB model on the DBs with a very large number of cases (in 11 out of the 14 large DBs, all having a multivalued class variable). This result seems to be consistent with the observation that the NB model, because of its large bias, can be considerably inaccurate even for very large sample sizes.

## 6    Conclusions and future work

We have presented a new Bayesian network classifier that combines the desirable properties of the NB classifier and the FM classifier, while relaxing the assumptions on which these models are based. We have also presented experimental results aimed at comparing the proposed model with the two components models, as well as at comparing four different techniques of model selection. These results provide evidence that the new classifier can often outperform the NB classifier in terms of classification accuracy, while significantly improving the calibration of the probability estimates.

The results are even more encouraging if we focus on those domains modeled by discrete variables only. As pointed out in the previous section, the use of unconstrained univariate Normal distributions could be the cause of the poor performance. A possible solution to this problem is to model the distribution of the continuous variables as a multivariate Normal with a common variance/covariance matrix, as is proposed in [20].

An extension of the model worth investigating is the use of hidden variables of different cardinality for different values of the class variable. That is, instead of selecting the hidden variable's cardinality based on the entire dataset, we can select a different cardinality for each subset corresponding to a given instantiation of the class variable. The resulting model would consist of a set of FM models, one for each value of the class variable, which could then be combined to do prediction with respect to the class variable.

## Acknowledgments

We thank Roger Day for his useful suggestions on issues of statistical testing, and the anonymous reviewers for their comments on a preliminary version of this manuscript. The research presented here was supported in part by grants IRI-9509792 and IIS-9812021 from the National Science Foundation and by grant LM05291 from the National Library of Medicine. Stefano Monti was supported in part by a Andrew Mellon graduate fellowship.